# covLLM: Large Language Models for COVID-19 Biomedical Literature

Yousuf A. Khan[*,1,2,3,4,5]　　　　　　　　　　　　　　　　　　Clarisse Hokia[*,1,6]

Jennifer Xu[*,6,7]　　　　　　　　　　　　　　　　　　　　　　　Ben Ehlert[*,1]

[*]All authors contributed equally to this work
[1]Department of Biomedical Data Science, Stanford University, CA, USA
[2]Department of Molecular and Cellular Physiology, Stanford University, Stanford University, CA, USA
[3]Department of Neurology and Neurological Sciences, Stanford University, Stanford, CA, USA
[4]Department of Structural Biology, Stanford University, Stanford, CA, USA
[5]Department of Photon Science, Stanford University, Stanford, CA, USA
[6]Department of Computer Science, Stanford University, Stanford, CA, USA
[7]Department of Bioengineering, Stanford University, Stanford, CA, USA

Emails: Yousuf@stanford.edu, chokia@stanford.edu, jennxu23@stanford.edu, behlert@stanford.edu

**Abstract**

The COVID-19 pandemic led to 1.1 million deaths in the United States, despite the explosion of coronavirus research. These new findings are slow to translate to clinical interventions, leading to poorer patient outcomes and unnecessary deaths. One reason is that clinicians, overwhelmed by patients, struggle to keep pace with the rate of new coronavirus literature. A potential solution is developing a tool for evaluating coronavirus literature using large language models (LLMs) – neural networks that are deployed for natural language processing. LLMs can be used to summarize and extract user-specified information. The greater availability and advancement of LLMs and pre-processed coronavirus literature databases provide the opportunity to assist clinicians in evaluating coronavirus literature through a coronavirus literature specific LLM (covLLM), a tool that directly takes an inputted research article and a user query to return an answer. Using the COVID-19 Open Research Dataset (CORD-19), we produced two datasets: (1) synCovid, which uses a combination of handwritten prompts and synthetic prompts generated using OpenAI, and (2) real abstracts, which contains abstract and title pairs. covLLM was trained with LLaMA 7B as a baseline model to produce three models trained on (1) the Alpaca and synCovid datasets, (2) the synCovid dataset, and (3) the synCovid and real abstract datasets. These models were evaluated by two human evaluators and ChatGPT. Results demonstrate that training covLLM on the synCovid and abstract pairs datasets performs competitively with ChatGPT and outperforms covLLM trained primarily using the Alpaca dataset.

# 1. Introduction

## 1.1. Covid-19

In just three years, over 103 million people in the United States tested positive for COVID-19 and over 1.1 million people in the United States died due to COVID-19 complications[1]. COVID-19 is a highly infectious viral disease caused by SARS-CoV-2. It can cause a wide range of symptoms, most commonly fever, chills, and sore throat. Depending on the severity of the symptoms, several patients require immediate medical attention for severe difficulty in breathing, confusion, chest pain, or other symptoms of severe illness. Additionally, certain populations with pre-existing health conditions, those over age 60, and unvaccinated individuals are at increased risk for severe illness, hospitalization, and death, though anyone can become sick with COVID-19. People infected with COVID-19 are also at risk of long COVID, which occurs when they experience prolonged fatigue, respiratory, and neurological symptoms[2].

Over the past 3.5 years, few COVID-19 treatments have been developed and refined. In addition to the COVID-19 vaccines[3,4], these include over-the-counter medication, prescription medication, and in-patient treatments. Healthcare providers may prescribe Paxlovid or Lagevrio for high risk individuals infected with COVID-19. Evusheld monoclonal antibodies are prescribed to immunocompromised individuals exposed to COVID-19. Patients with severe illness due to COVID-19 who require hospitalization may be treated with the antiviral medication remdesivir and medications to counteract overactive immune systems or to treat complications. Through clinical trials and other research, these treatments were eventually developed and made available to both adult and pediatric patients who qualify. The NIH's Accelerating COVID-19 Therapeutic Interventions and Vaccines (ACTIV) initiative has promoted additional research on treatments such as immune modulators, monoclonal and polyclonal antibodies, and blood thinners and on the uses of medications used to treat other conditions[5]. However, this research has been slow to translate to clinical treatments.

Vaccine and treatment developments for COVID-19 were developed through an acceleration in research and treatments. Between January 1, 2020 and June 30, 2020, researchers published over 23,500 coronavirus articles, letters, reviews, notes, and editorials to major databases[6]. By August 1, 2021, the number of publications increased to 210,183, with 720,801 unique authors from all scientific subfields[7]. This vast involvement of the scientific research community was unlike trends from other infectious diseases, including HIV/AIDS, Zika, and tuberculosis. The United States, China, and Italy were the countries that published the most papers by volume, while BMJ, Journal of Medical Virology, and The Lancet were the journals that published the most papers by volume. Of the articles published on Scopus and Web of Science, 48% and 37%, respectively, were research papers. Findings have involved topics such as data reporting quality, mental health impacts of the pandemic, conflicts of interest, quality of research publications and studies, impacts of the pandemic on academia, and the uses of technology to learn more about COVID-19[6]. Additionally, at least one author from each of the 21 major scientific fields and 174 scientific subfields published on COVID-19[7].

Through the race to publish on the COVID-19 pandemic, scientists have highlighted the volume of articles and questioned the quality of clinical trials. Ioannidis, Salholz-Hillel, et. al underscore the number of researchers and breadth of disciplines that published on COVID-19, stating that 28% of COVID-19 publication authors published in a subfield that was different from their subfield of expertise. They express concern that some COVID-19 authors' fields of expertise were "remote" from COVID-19, including "fisheries, ornithology, entomology, or architecture". They also cite that some scientists had participated in "epistemic trespassing," where scientists publish on health and medical questions, despite being experts in other fields. Moreover, surveys on the quality of COVID-19 research consistently found a high prevalence of low-quality studies[7]. Park, Mogg, et. al argue that clinical trials focused primarily on treatments for severe disease, rather than pre-exposure, post-exposure, or outpatient treatments and identified shortcomings including overlaps in proposed trials, having sample sizes smaller than 100 participants, and not identifying dose ranges. The translation of relevant findings to clinical practice has been slow and inconsistent, resulting in poorer quality of care to patients[8]. This exponential rise in coronavirus research creates the opportunity for computational methods that enable clinicians to efficiently filter through papers and rapidly translate these findings into treatments.

### 1.2. Large Language Models

Large language models (LLMs) are deep learning algorithms that can engage with linguistic and language components, such as text, for natural language processing and other artificial intelligence applications. LLMs learn from large datasets that typically include almost everything available on the internet, where algorithms define metrics of similarity and use those to group inputs. Following training, LLMs are then able to use the given knowledge to generate desired outputs. They can also be trained or fine-tuned with smaller batches of data for specific applications, such as biomedical research. One commonly known example of an LLM application is ChatGPT, which can perform natural language processing functions. Despite the potential applications of LLMs, some challenges of using LLMs for specific fields include domain constraints, dataset availability, and technical skillset of the developers[9]. Branching from current technologies, emerging areas of LLMs include developing models that can check their own outputs. A current shortcoming of existing generative language models is models' tendencies to "hallucinate", which occurs when LLMs present false or inaccurate information as facts. Potential solutions to this problem include having a model provide citations, allowing the model to access external information sources, or asking the model to identify aspects of the output that it feels are the weakest[10].

In addition to the development of powerful large language models, several LLMs have also been made available to researchers. For example, in February 2023, Meta publicly released Large Language Model Meta AI (LLaMA), an LLM that works with less compute and resources and provides researchers access to studying and fine-tuning LLMs. This allows researchers to further understand how LLMs work, to improve LLMs, and to reduce issues like bias and misinformation[11]. One application of LLaMA is Stanford Alpaca, a fine-tuned model that can behave similarly to OpenAI's text-davinci-003 and follow instructions. However, due to ethical issues, safety concerns, and companies' policies, both are only available for academic research, and LLaMA is released under a non-commercial license[12].

Opportunities that make the creation of a COVID-19-specific LLM are the availability of biomedical literature datasets and machine learning applications that process COVID-19 literature. For instance, in June 2020, the Machine Learning Google Developer Experts group (ML GDEs) released the first version of the Biomedical Research Extensive Archive To Help Everyone (BREATHE), which is a large-scale database with over 16 million biomedical articles from different repositories and hosted on Google BigQuery. This publicly accessible database contains titles, abstracts, and some full body texts and allows biomedical researchers to glean new insights from research publications[13]. Additionally, the CoronaCentral resource used a BERT-based document multilabel classification method to categorize nearly 130,000 papers based on topic, article type, and preprint server type so that users can identify papers that are most pertinent to their research or clinical needs[14]. A biomedical LLM, known as BioMedLM, has also been released by the Stanford Center for Research on Foundation Models (CRFM) and MosaicML. This model was trained on biomedical data from PubMed and demonstrated that training LLMs on data from specific fields can outperform general-purpose models. Other LLMs created by the CRFM team include DRAGON and BioLinkBERT. Work by the CRFM team has shown that LLMs are applicable to specific fields and that focusing the model on a specific field allows models to perform well with less data and compute[15].

With the availability of foundational and fine-tuned LLMs, biomedical literature databases, and prior work on COVID-19 literature classification, we fine-tuned a large language model to interact with COVID-19 literature inputs and queries called covLLM.

## 2. Methods

### 2.1. Description of Data

We generated two types of training data: a) synthetic training data generated through OpenAI's text-davinci-003 model with diverse prompts and content and b) actual abstracts where the only provided prompt was to summarize the abstract and the output was the actual title of the article.

The BREATHE dataset was used for generating both types of training data, serving as the basis of synthetic data generation (described below) or for mining of real abstracts. BREATHE is a large biomedical literature database containing papers from 10 major repositories of biomedical research. We specifically sample from CORD-19, a subset of BREATHE that contains curated articles deemed relevant to COVID-19 research[16,17].

#### 2.1.1. Synthetic Data Generation

To generate our training data, we followed the self-instruction protocol[18.]. In line with the self-instruction input format, our training dataset is a list of instruction-input-output triplets (**Figure 1**). To create the initial seed tasks, we paired 18 handwritten instructions with 175 randomly selected abstracts from the CORD-19 dataset. Examples of possible instructions include summarizing a provided abstract, extracting the key findings, identifying any mentioned biological or chemical pathways, determining the study type, and evaluating the quality of a study's findings. Using OpenAI's gpt-turbo-3.5 model, we utilized the instruction-abstract pairs to generate the corresponding outputs. Each output was manually evaluated and edited for comprehensibility, correctness, and conciseness.

We then employed Alpaca's self-instruction-based data synthesis pipeline[19] to generate a total of 1097 instruction-input-output triplets. The pipeline utilizes a directed prompt and OpenAI's text-davinci-003 to generate synthetic instruction-input-output triplets from a given set of seed tasks. We modified Alpaca's directed prompt to guide synthetic tasks towards biomedical research-related topics, ensuring each task included an input formatted as a 250-300 word abstract.

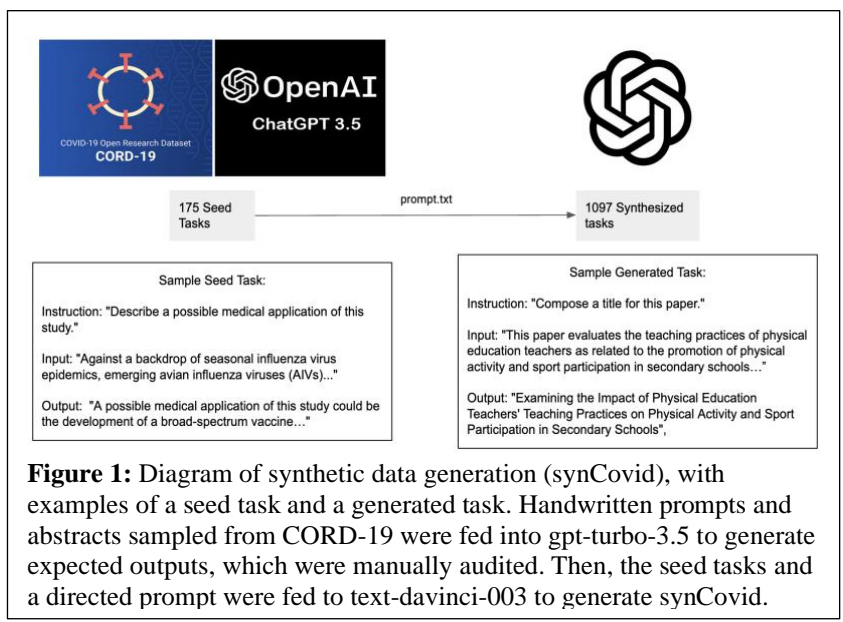

**Figure 1:** Diagram of synthetic data generation (synCovid), with examples of a seed task and a generated task. Handwritten prompts and abstracts sampled from CORD-19 were fed into gpt-turbo-3.5 to generate expected outputs, which were manually audited. Then, the seed tasks and a directed prompt were fed to text-davinci-003 to generate synCovid.

We chose to generate a small training set size of 1097 examples, as compared to Alpacas' 52,000 training set, due to the monetary cost of generating these examples and based on the results of another study that demonstrated training on as little as 1000 instructions can yield robust performance if properly fine-tuned[20]. This collection of synthetic COVID19 instructions, synCovid, consists of 1097 instruction-input-output triplets.

### 2.1.2. Real abstract mining

In addition to the synthetic training set, we also created a simple dataset of instruction-input-output triplets in which the inputs were real, COVID19 specific abstracts. For each entry in this dataset, the instruction is "Summarize this abstract", the input is an abstract sampled from CORD-19, and the output is the actual title associated with the selected abstract. We sampled 1097 examples for this training data, equal to the number of synthetically generated instructions.

## 2.2. Developing our models

### 2.2.1. Training covLLM

We trained covLLM using the LLaMa 7B model as our baseline model[21]. We ultimately trained **three different models** (the classic Alpaca 52K self-instruction dataset supplemented with 1097 synthetic scientific literature specific tasks, 1097 synthetically generated tasks, and 1097 input, real abstract paired prompts). We fine-tuned our models using the Alpaca-Lora framework[22,23], which only required several hours on a single NVIDIA A100 per model. The relevant training parameters for the following datasets were the following: 1) Alpaca 52K + synCovid dataset – 53097 total instructions, 3 epochs, learning rate of 3e-4, batch size of 128, and eval size of 2,000 2) synCovid dataset only – 1097 total instructions, 30 epochs, learning rate of 1e-5, batch size of 16, eval size of 100 3) synCovid and real abstract paired prompts – 2194 instructions, 30 epochs, learning rate of 1e-5, batch size of 16, eval size of 100. These parameters were determined by a parameter sweep and by assessing the training and evaluation loss curves (data not shown). Otherwise, all other parameters were kept identical to the Alpaca-Lora framework.

### 2.3. Evaluating our models

#### 2.3.1. Experiment

We conducted an experiment to assess the performance of our three models, namely synCovid, synCovid+abstracts and synCovid+Alpaca against ChatGPT in generating satisfactory outputs. The purpose was to evaluate how well these models can respond to various test prompts.

We devised an experimental setup where we generate a single response to each test prompt from each model. Responses were blinded to the human evaluators and ordering was randomized. Two Human evaluators compared the responses and indicated their preference for each prompt. We also repeated the experiment using GPT-3.5 as the evaluator.

#### 2.3.2. Model Output Generation

For generating outputs from our model for test set evaluation (i.e. inference), we used the following parameters for all three major models that we trained: Temperature: 0.1, Top p: 0.75, Top k: 40, Beams: 4, and Max Tokens: 128. For generating ChatGPT outputs to compare these models against, we simply prompted it with the following input: "Please respond to these instructions with a given input in a few sentences; assume that each question is independent of each other and answer each one individually."

#### 2.3.3. Methodology

During each iteration, the evaluators received the instruction (e.g. "Summarize this abstract"), input (e.g. text of the abstract) and 4 responses from the models (**Figure 2**). The evaluators were requested to rank each model considering helpfulness, relevance, accuracy, and level of detail, and ties between models were allowed. Furthermore, the evaluators scored each model as either **Fail**: the response did not meet the requirements of the prompt, **Pass**: the response met the requirements of the prompt, or **Excellent**: the model provided an excellent response to the prompt. This follows the same grading system as described in the LIMA study[20]. The specific prompt for GPT3.5 evaluation is in the appendix.

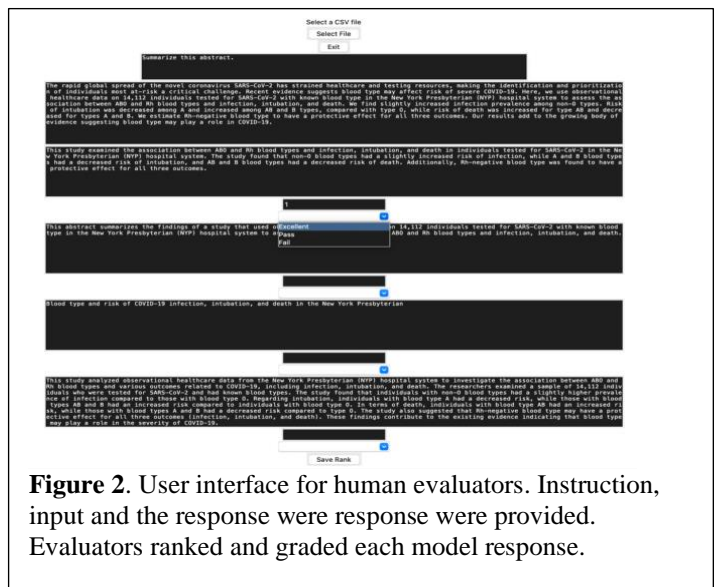

**Figure 2**. User interface for human evaluators. Instruction, input and the response were response were provided. Evaluators ranked and graded each model response.

To evaluate each model, we counted the number of **Excellent**, **Pass**, and **Fail** grades then averaged the results from the three sets of evaluations. This was repeated for the models' rankings from 1 to 4. Results from the two human evaluators and GPT3.5 were weighted equally, and the models' training dataset(s) remained blinded until evaluations were completed.

## 3. Results

### 3.1. Data Generation

#### 3.1.1. Synthetic Data Generation Quality Control

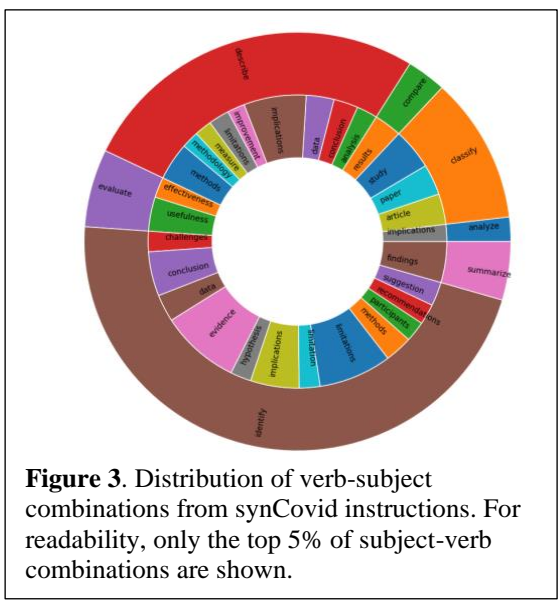

**Figure 3**. Distribution of verb-subject combinations from synCovid instructions. For readability, only the top 5% of subject-verb combinations are shown.

We devised a synCovid, a synthetic data generated by OpenAI's text-davinci-003 model, as one of our sources for training data. synCovid is a dataset of instruction-input-output triplets consisting of 1035 unique instructions and 865 unique inputs for a total of 1097 aggregate instructions. To evaluate the diversity and quality of synCovid prior to its inclusion into training, we examined the generated instructions in two ways.

First, we manually extracted verb-subject pairs from each of the synCovid instructions, resulting in 581 unique verb-subject pairs (**Figure 3**). The majority of the instructions were related to extracting specific information from the given input, such as identifying the sample population or describing the study methodology.

The diversity of the synCovid instructions can be seen through the subjects, which are more equally represented in the verb-subject pairs.

To assess the quality of the synCovid generated inputs, we randomly sampled 120 synCovid examples. Ideally, these inputs mimic an abstract of a biomedical research paper. Therefore, an input was considered complete if it discussed background information, methodology, results, and conclusions. We classified each sample input as complete or incomplete. We also determined the study design described by input. Our sampled generated inputs are representative of a variety of study designs. The most common study designs generated were literature reviews, cross-sectional studies, and method development studies (**Figure 4**). While all the sampled generated inputs were comprehensible, a minority were incomplete. Some generated abstracts consisted solely of a methodology and description of results (**Figure 5**). Despite this, we decided to include both fully complete and partially complete in our training data due to the prompt diversity they provided.

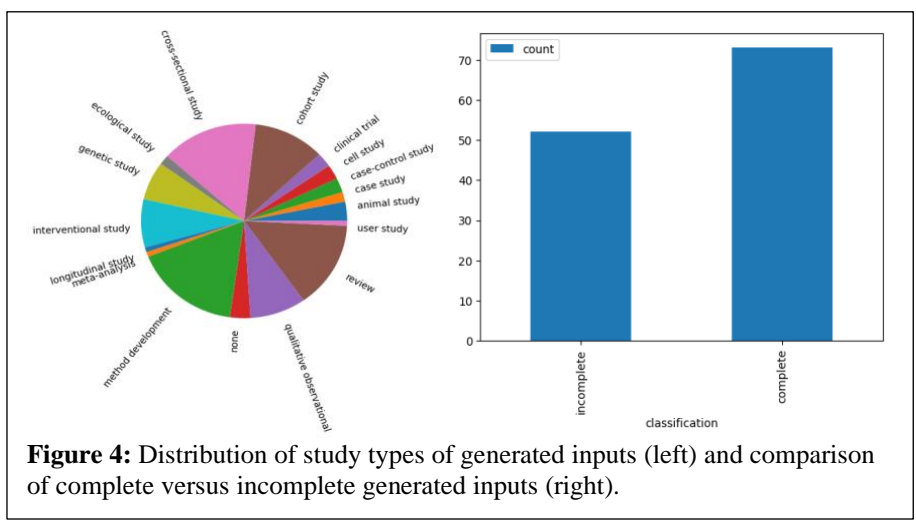

**Figure 4:** Distribution of study types of generated inputs (left) and comparison of complete versus incomplete generated inputs (right).

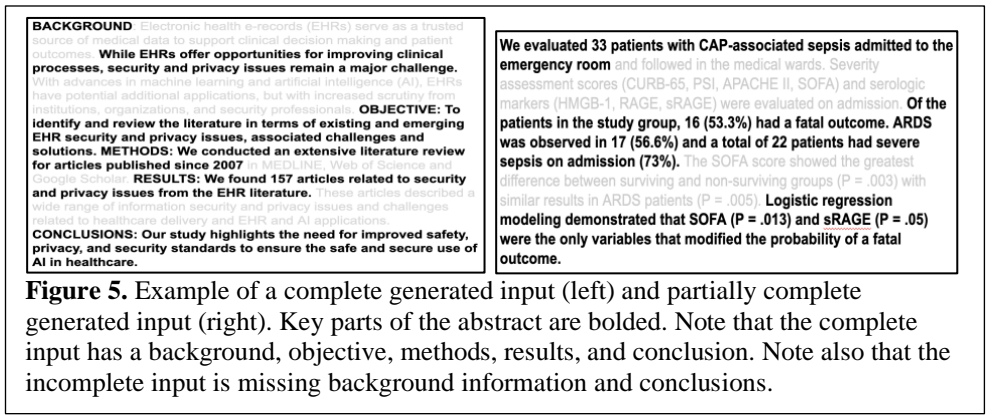

**Figure 5.** Example of a complete generated input (left) and partially complete generated input (right). Key parts of the abstract are bolded. Note that the complete input has a background, objective, methods, results, and conclusion. Note also that the incomplete input is missing background information and conclusions.

### 3.1.2. Synthetic Data Generation Quality Control

We manually double checked the results of our real abstract mining to ensure that our abstract instructions were of good quality.

## 3.2. Developing well-trained models

### 3.2.1. covLLM Training Results

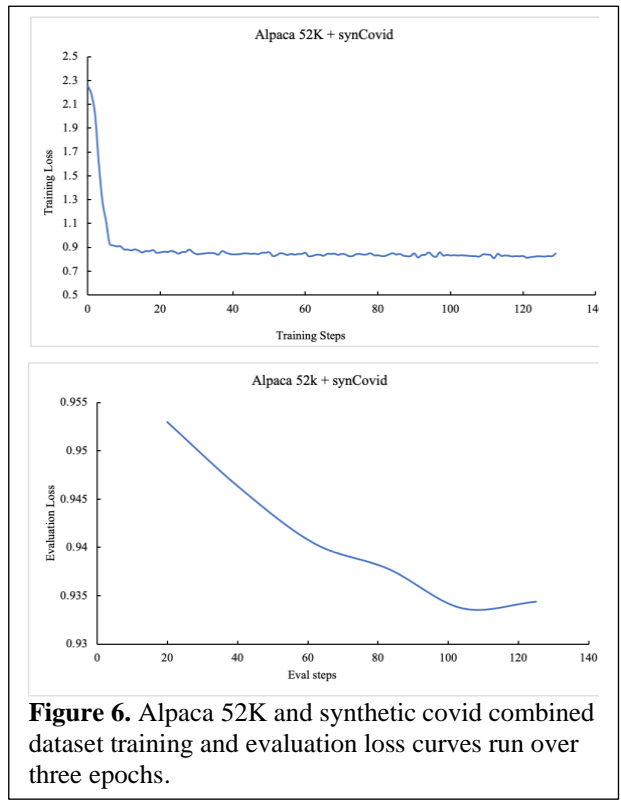

**Figure 6.** Alpaca 52K and synthetic covid combined dataset training and evaluation loss curves run over three epochs.

Here we present our training and evaluation curves for our three major models after training. As expected, the Alpaca + synCovid model showed both a decrease in training and evaluation loss over the course of our training, demonstrating that the model was not overfit (**Figure 6**). Overfitting was a major concern we had using such small training sets for our synCovid only (1097 unique instructions) and synCovid + real abstract prompts (2194 instructions). However, our training and evaluation curves demonstrate that, despite cycling through these limited datasets for 30 epochs, we did not overfit our model (**Figure 7**).

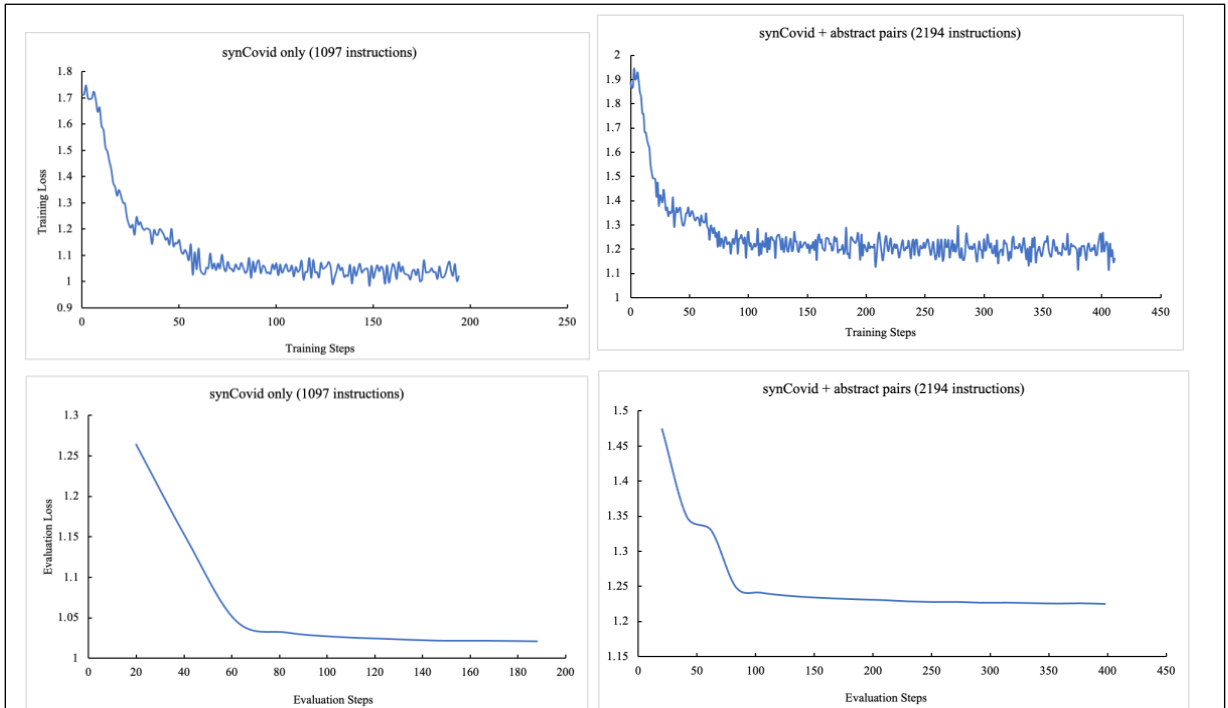

**Figure 7.** The two left panels are from the synthetic covid19 dataset only (1097 instructions) and the right two panels are from the synthetic covid19 dataset and the real abstract pairs database (2194 instructions). Both regimens show a decrease in training and evaluation loss over 30 epochs of data.

## 3.3. How the models performed

### 3.3.1. – 3.3.4. Evaluation summary

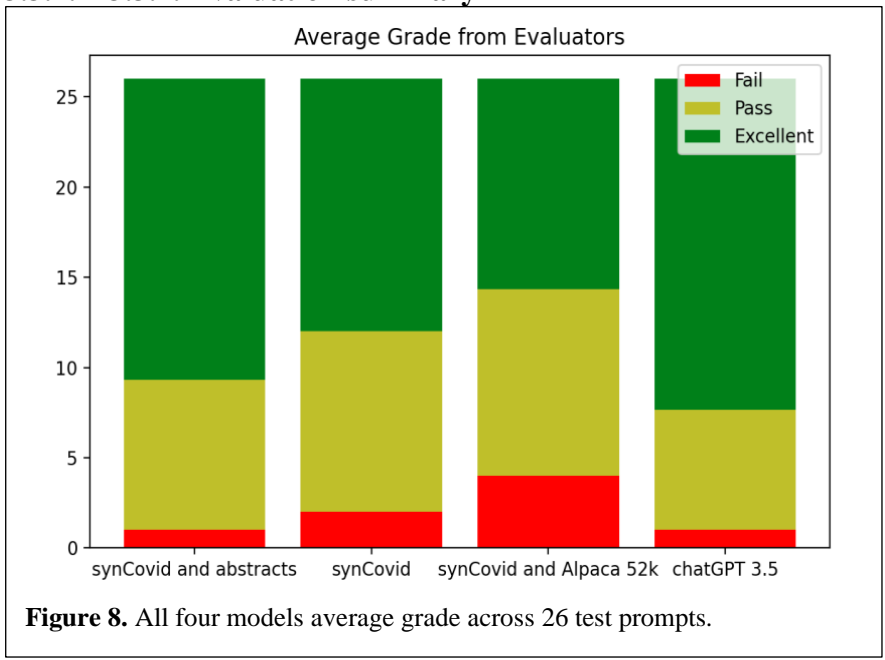

**Figure 8.** All four models average grade across 26 test prompts.

In this section, we present the findings from our evaluation, where the evaluators assessed a total of 26 instruction-input pairs. **Figure 8** illustrates the average grade assigned to the models across all evaluators. Several key observations emerged from the evaluation process. First, we observed that the inclusion of abstract-summarization pairs proved to be beneficial. The model that incorporated these pairs (synCovid+abstracts) showed great performance. Second, we discovered that the addition of the 52k Alpaca tasks to the synCovid model did not lead to any significant enhancement in performance.

In **Figure 9**, we present a head-to-head comparison between our models and ChatGPT. Notably, we found that the synCovid+abstracts model emerged as the best performing model, exhibiting promising results. In 65% of the prompts, this model was either preferred by the evaluators or tied with ChatGPT in terms of performance.

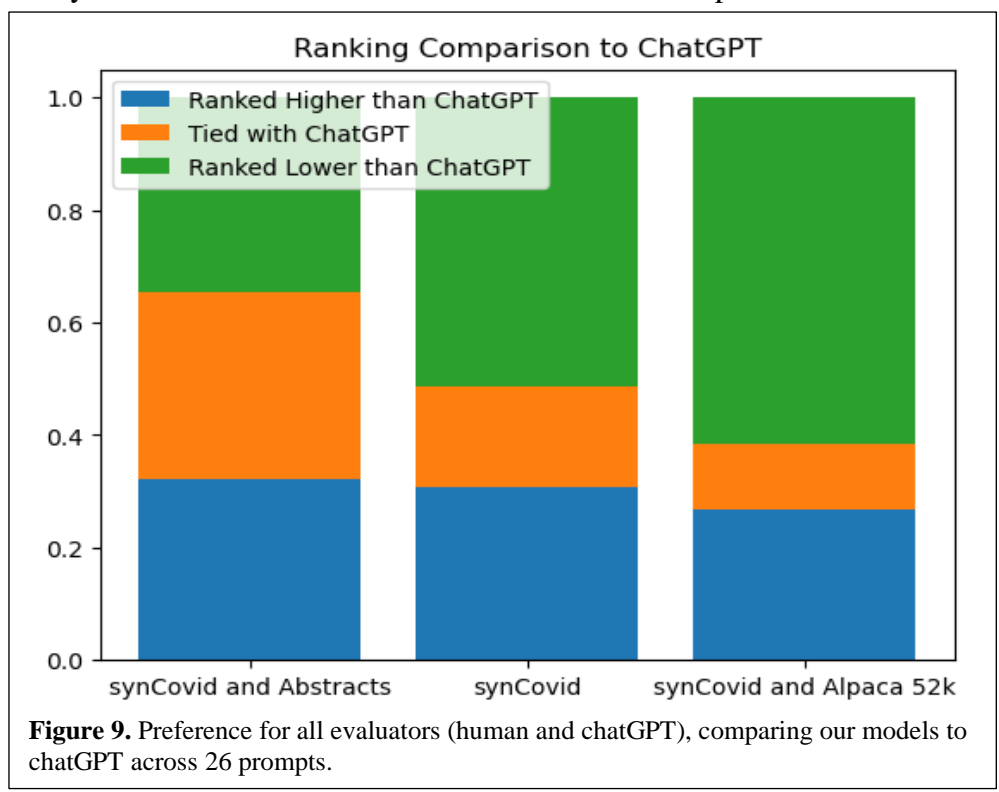

**Figure 9.** Preference for all evaluators (human and chatGPT), comparing our models to chatGPT across 26 prompts.

## 4. Discussion

The COVID-19 pandemic simultaneously led to millions of deaths globally and a need to rapidly translate research to treatment. covLLM, a machine learning-based tool that we successfully developed, will enable scientists and clinicians to rapidly incorporate knowledge from the growing body of literature into their decisions, research, and clinical care that will impact patient outcomes. covLLM will provide an architecture to tackle current diseases and future pandemics. Additionally, pilot studies into other research fields (data not shown) demonstrates that covLLM's base architecture and training strategy can be generalized to additional scientific field.

From an LLM perspective, we demonstrate how a general, small pre-trained language model can be guided and fine-tuned to accomplish a highly specific task given a handful of synthetically generated and mined tasks. This was especially surprising, given that the synthetically generated instructions and abstracts contained hallucinated information but were still matched the style and syntax of biomedical research. Additionally, covLLM performance was comparable or exceeded ChatGPT's performance in our evaluation. This emphasizes the importance of prompt diversity, not necessarily prompt accuracy, in the fine-tuning stage and shows how limited real-world data can still lead to robust performance.

      The limitations of covLLM to practical usage by clinicians and researchers is two-fold. First, to quickly process entire papers in a reasonable time frame, a powerful and dedicated machine must be setup to take such requests. This is currently impractical given our current limitations as students, but could be solved if given additional resources and time to optimize the performance of these models on more consumer-based hardware. Another limitation is that its answers to broader, more philosophical questions such as "What are the public health implications of this basic science research?" can range from highly accurate to inaccurate. Thus, some degree of user filtering and prior knowledge is still required, as with many other computational tools and assistants.

## 5. Appendix

See https://github.com/clarisseh47/bioLLM.